\documentclass{article}

\usepackage{arxiv}

\usepackage[utf8]{inputenc} % allow utf-8 input
\usepackage[T1]{fontenc}    % use 8-bit T1 fonts
\usepackage{hyperref}       % hyperlinks
\usepackage{url}            % simple URL typesetting
\usepackage{booktabs}       % professional-quality tables
\usepackage{amsfonts}       % blackboard math symbols
\usepackage{nicefrac}       % compact symbols for 1/2, etc.
\usepackage{microtype}      % microtypography
\usepackage{lipsum}		% Can be removed after putting your text content
\usepackage{graphicx}
\usepackage{doi}
\usepackage{amsmath,amsfonts}
\usepackage{algorithmic}
\usepackage{xcolor}

\usepackage{algorithm}
\usepackage{array}
\usepackage[caption=false,font=normalsize,labelfont=sf,textfont=sf]{subfig}
\usepackage{textcomp}
\usepackage{stfloats}
\usepackage{url}
\usepackage{verbatim}
\usepackage{graphicx}
\usepackage{cite}

\title{Automated Linear Disturbance Mapping via Semantic Segmentation of Sentinel-2 Imagery}

\author{Andrew M. Nagel \\
	Department of Computer Science\\
	University of Winnipeg \\
	Winnipeg, MB \\
	\texttt{a.nagel@uwinnipeg.ca} \\
	\And
    Anne Webster\\
	Hatfield Consultants Group \\
    Geomatics and Remote Sensing \\
	North Vancouver, BC, CA \\
	\texttt{awebster@hatfieldgroup.com} \\
    \And
    Christopher Henry \\
    Department of Computer Science \\
    University of Manitoba \\
    Winnipeg, MB, CA \\
    \texttt{christopher.henry@umanitoba.ca}
	\AND
	Christopher Storie \\
	Department of Geography \\
	University of Winnipeg \\
    Winnipeg, MB, CA
	\texttt{c.storie@uwinnipeg.ca} \\
	\And
	Ignacio San-Miguel Sanchez \\
	Pachama \\
	San Francisco, California, USA \\
	\texttt{ignaciosanmiguel86@gmail.com} \\
	\AND
	Olivier Tsui \\
	Hatfield Consultants Group \\
    Geomatics and Remote Sensing \\
    North Vancouver, BC, CA \\
	\texttt{otsui@hatfieldgroup.com} \\
    \And
    Jason Duffe \\
    Environment and Climate Change Canada \\
    Landscape Science and Technology \\
    Gatineau, QC, CA \\
    \texttt{Jason.Duffe@ec.gc.ca} \\
    \And
    Andy Dean \\
    Hatfield Consultants Group \\
    North Vancouver, BC, CA \\
    \texttt{adean@hatfieldgroup.com}
}

\begin{document}
\maketitle

\begin{abstract}
In Canada's northern regions, linear disturbances such as roads, seismic exploration lines, and pipelines pose a significant threat to the boreal woodland caribou population (Rangifer tarandus). To address the critical need for management of these disturbances, there is a strong emphasis on developing mapping approaches that accurately identify forest habitat fragmentation. The traditional approach is manually generating maps, which is time-consuming and lacks the capability for frequent updates. Instead, applying deep learning methods to multispectral satellite imagery offers a cost-effective solution for automated and regularly updated map production. Deep learning models have shown promise in extracting paved roads in urban environments when paired with high-resolution (\textless~0.5m) imagery, but their effectiveness for general linear feature extraction in forested areas from lower resolution imagery remains underexplored. This research employs a deep convolutional neural network model based on the VGGNet16 architecture for semantic segmentation of lower resolution (10m) Sentinel-2 satellite imagery, creating precise multi-class linear disturbance maps. The model is trained using ground-truth label maps sourced from the freely available Alberta Institute of Biodiversity Monitoring Human Footprint dataset, specifically targeting the Boreal and Taiga Plains ecozones in Alberta, Canada. Despite challenges in segmenting lower resolution imagery, particularly for thin linear disturbances like seismic exploration lines that can exhibit a width of 1-3 pixels in Sentinel-2 imagery, our results demonstrate the effectiveness of the VGGNet model for accurate linear disturbance retrieval. By leveraging the freely available Sentinel-2 imagery, this work advances cost-effective automated mapping techniques for identifying and monitoring linear disturbance fragmentation.
\end{abstract}

% keywords can be removed

\section{Introduction}

In the Canadian boreal forest region, habitat fragmentation caused by linear disturbances (LDs) such as roads, seismic exploration lines (cutlines), pipelines, and energy transmission corridors is a leading factor in the decline of woodland caribou (Rangifer tarandus), particularly in western Canada \cite{ecc2020}. 
Consequently, understanding the extent, spatial distribution, and dynamics of these disturbances has become a critical priority for research and forest management in Canada.
Under the Species at Risk Act, Canada has imposed regulatory restrictions on the density of forest habitat disturbances within woodland caribou ranges \cite{ecc2020}. 

To support this regulation, government agencies currently rely on manual digitization of LDs using satellite imagery over vast areas. 
Examples of these datasets include the Anthropogenic Disturbance Footprint Canada, which maps LDs for over 51 priority herds across millions of hectares using visual interpretation of Landsat data for 2008-2010 at 30m resolution and for 2015 at both 30m and 15m using the panchromatic band \cite{adfc}; and the Human Footprint (HF) Wall-to-Wall dataset, a vector polygon layer capturing disturbances across Alberta from 2010 to 2021 \cite{abmi}.
While these efforts provide valuable LD data, they are time-consuming and expensive, resulting in incomplete and infrequent coverage. 
Therefore, cost-effective methods for periodically mapping LDs in forest settings across Canada are needed to better support management decisions and conservation efforts at large scales.

Automated techniques employing machine learning offer an appealing solution for consistently mapping LDs across vast territories while minimizing costs. 
This methodology revolves around semantic segmentation, wherein each pixel in an image is assigned a class label (e.g., ``road", ``pipeline", ``cutline", or ``background"). 
While traditional machine learning methods like Random Forest or Support Vector Machines can perform this task \cite{rf, svm}, they often fall short in capturing the intricate spatial context crucial for accurate classification of LD features. 
Deep learning (DL) convolutional neural networks (CNNs), however, excel in this domain by explicitly incorporating neighboring pixel information through convolutional operations to enable the modelling of increasingly long-range spatial dependencies.

In 2015, Long et al. proposed an encoder-decoder architecture for semantic segmentation \cite{long2015}. 
This architecture utilizes convolutional neural networks (CNNs) to extract features from the input imagery through an encoder, which are subsequently decoded into segmentation maps using a reversed CNN. 
Over time, this architecture has been refined by modifying the encoder and/or decoder structure \cite{simonyan2015, deeplab}.
For instance, the U-Net architecture, originally developed for biomedical imaging, introduced skip connections that link the encoder and decoder layers, preserving spatial information throughout the network \cite{unet2015, deeplab}.

The adoption of encoder-decoder networks in remote sensing applications has achieved notable success, delivering precise land cover and land use maps \cite{henry2019, alhassan2020, tsenov2023}.
Moreover, automatic extraction of roads from satellite imagery using DL has gained increasing attention in recent years \cite{chen2022, cheng2017, ding2020}. 
However, many of these methods focus on detecting urban roads.
In contrast, extracting LDs in forest areas, which vary drastically in appearance, is notably more complex than mapping urban paved roads.
For example, roads in forested regions differ significantly in surface conditions and surrounding vegetation cover, making them harder to predict compared to the more homogeneous urban roads.
Additionally, forested areas contain various types of LDs, such as cutlines, pipeline corridors, and logging roads, each with unique characteristics. 
Unlike urban roads, these features are narrower, more irregular, and often obscured by vegetation regrowth, further complicating their detection \cite{calsys2023}. 

While a few publications have explored forest/logging road extraction using DL from satellite data in Canada and elsewhere \cite{trier2022,zhang2021}, none have focused on capturing multiple LD classes in forested areas. 
A significant hurdle is the resolution of imagery, as many existing techniques have been developed using very high-resolution imagery (\textless~0.5m) and have not been adapted for freely available lower-resolution imagery, such as the 10m Sentinel-2 (S2) data.
Therefore, the objective of this research is to develop and assess the accuracy of an automated DL approach for extracting general LDs in forested areas using lower-resolution S2 imagery.

\section{Methodology}

\subsection{Data and Preprocessing}

\begin{figure}[ht]
  \centering
  \includegraphics[width=0.4\textwidth]{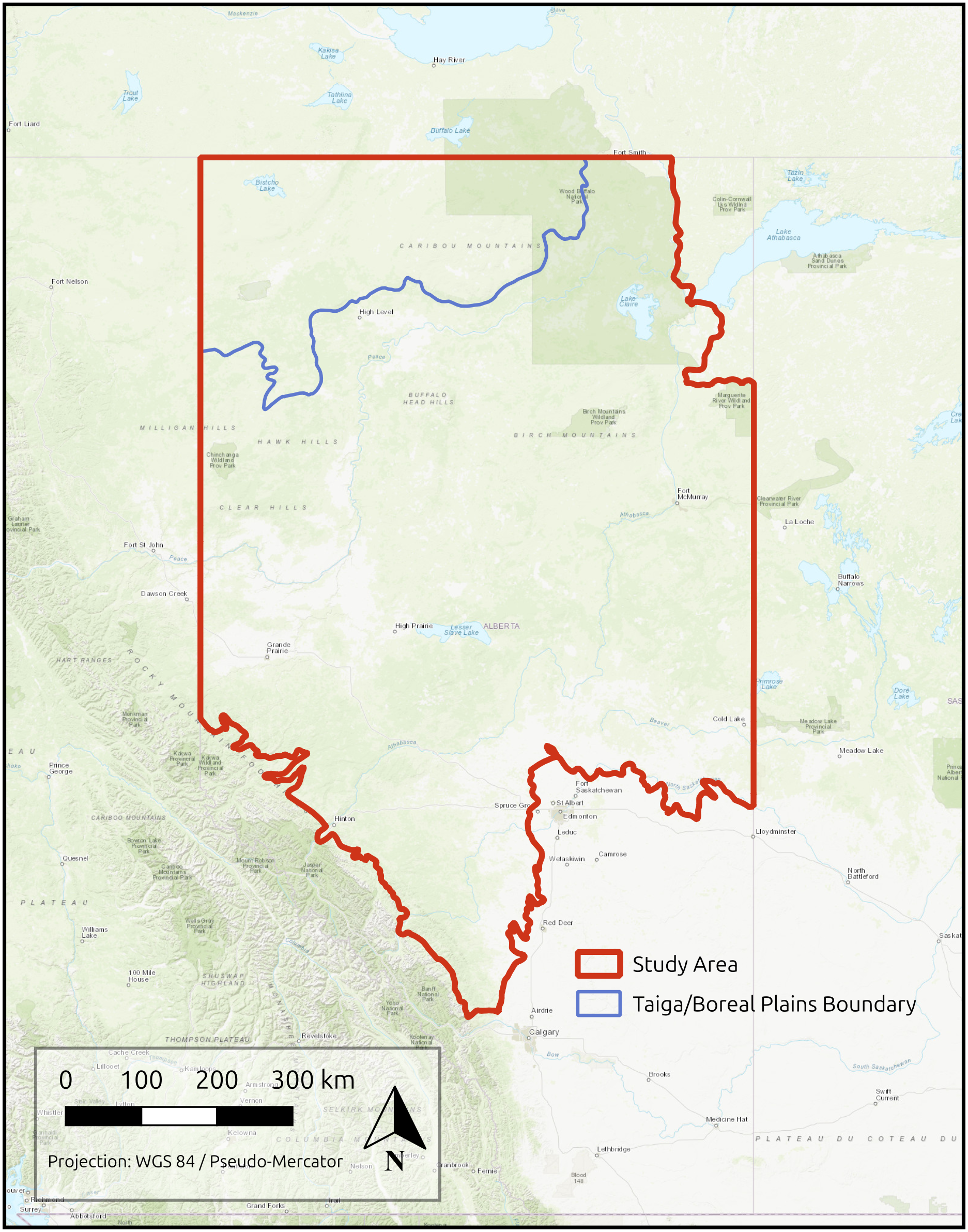}
  \caption{Study area in Alberta, Canada (denoted in red), encompassing the Boreal Plains (below the blue boundary) and Taiga Plains ecozones (above the blue boundary).}
  \label{fig:plains_ecozones}
\end{figure}

This study focuses on a specific region of Alberta, Canada, encompassing the Boreal and Taiga Plains ecozones (as shown in Fig.~\ref{fig:plains_ecozones}). 
The study area spans 44.32 million hectares and encompasses a variety of landscapes, including urban, agricultural, rural, and forested regions.
The area is characterized by a high density of linear disturbances, including those affecting Caribou habitat, and has extensive relevant datasets available. 
S2 satellite imagery was used as the input predictor variable, while the HF vector dataset provided the ground truth labels for the study. 
Detailed descriptions of these datasets are provided in the following sections.

\subsubsection{Sentinel-2 Imagery}

Data from the S2 mission, part of the European Union (EU) Copernicus Program, are freely available under an open license, making it an excellent candidate for cost-effective wide-area mapping.
Moreover, S2 data offers the highest resolution available among free satellite data sources.
S2 is a constellation of two satellites (A \& B) that collect multispectral information with a spatial resolution of 10 and 20 meters, depending on the specific spectral bands. 
Sensors collect information in the visible, near-infrared, and short-wave portions of the electromagnetic spectrum. 

The 10-meter resolution S2 imagery utilized in this study was sourced from the Google Earth Engine platform's S2\_SR\_HARMONIZED collection to leverage the Cloud Score+ cloud masking algorithm. 
All S2 granules covering the Taiga and Boreal Plains ecozones of Alberta, Canada, were selected, and cloud effects were masked out using Cloud Score+ in combination with the cs\_cdf quality assessment band, with a clear threshold parameter set at 0.65. Following this, a median composite was created spanning the period from May 1, 2021, to July 31, 2021.
For this study, all 10m bands were retained, including the red, green, blue, and near-infrared (RGBNir) bands.

\subsubsection{Human Footprint Linear Disturbance Labels}

The ground truth label data for our study was obtained from the Human Footprint (HF) Wall-to-Wall dataset, developed by the Alberta Biodiversity Monitoring Institute (ABMI) \cite{abmi}. 
This dataset consists of a vector polygon layer that meticulously maps anthropogenic disturbances across Alberta. 
Updated annually from 2010 to 2021, it offers a comprehensive view of human impacts on the landscape.
The HF dataset was created by combining existing datasets, including Alberta Base Features, Inventories, and Road/Railway Networks. 
This merged dataset was then refined through visual interpretation of 1.5m-resolution SPOT-6 imagery.

ABMI defines the human footprint as ``the visible alteration or conversion of native ecosystems to temporary or permanent residential, recreational, agricultural, or industrial landscapes." 
This definition encompasses all areas under human use that have undergone significant changes from their natural state for prolonged periods. 
Such alterations include urban areas, roads, agricultural fields, and surface mines. 
Additionally, it encompasses land periodically disturbed by industrial activities, such as forestry cut blocks and cutlines (seismic exploration lines). 
Notably, some human land uses, like grazing, hunting, and trapping, are not currently factored into the human footprint analyses. 
Overall, the HF dataset is comprised of over 100 thematic classes.

\begin{figure*}[t]
  \centering
  \includegraphics[width=1.0\textwidth]{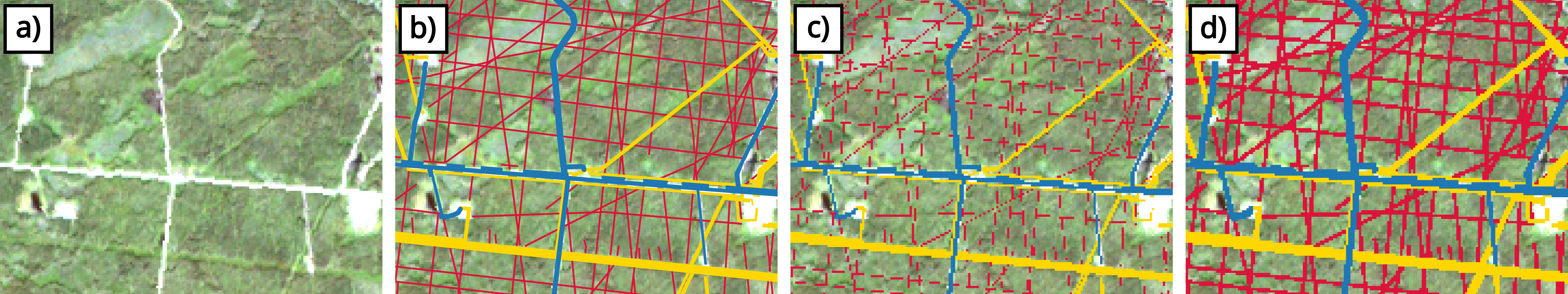}
  \caption{Visualization of Sentinel 2 imagery and ABMI linear disturbance labels. The color scheme is as follows: blue for roads, yellow for pipelines, and red for cutlines. \textbf{a)} Original Sentinel-2 imagery. \textbf{b)} Sentinel-2 imagery overlaid with ground-truth ABMI vector labels. \textbf{c)} Sentinel-2 imagery with ABMI labels rasterized based on features touching the pixel center. \textbf{d)} Sentinel-2 imagery with ABMI labels rasterized when features touch any part of the pixels.}
  \label{fig:cutline_raster}
\end{figure*}

Notably, roads, pipelines, and cutlines constitute 96\% of the LDs in the HF dataset, while other disturbances such as electrical transmission lines and railways are infrequent. 
Therefore, this study focused solely on pipelines, roads, and cutlines as the primary classes of interest. 
Additionally, although the HF dataset delineates 14 distinct classes of ``road," they were treated collectively as a single class for the purposes of this study.
An example of S2 imagery with ABMI vector linear disturbance labels is shown in Fig.~\ref{fig:cutline_raster}b).

The HF dataset was rasterized using the S2 data as a template to align the datasets. 
Pixels were assigned to the labeled data class that touched any portion of the pixel. 
For thinner disturbances like cutlines, this rasterization approach results in artificially thicker features as shown in Fig.~\ref{fig:cutline_raster}d).
Note that this effect will negatively impact the testing performance of the segmentation model, making it appear worse than it actually is, when trained on these thicker labels.
However, this rasterization method was still preferred over assigning a pixel class only if the vector label intersects with the pixel center.
The latter approach caused many narrow cutlines to become fragmented as shown in Fig.~\ref{fig:cutline_raster}c), leading to its own set of training challenges and validation errors.

\subsubsection{Tiling}

The S2 imagery and the corresponding rasterized HF labels were tiled into non-overlapping $224 \times 224$ pixel tile pairs. 
A tile pair was discarded if the image was \textless~5\% covered or the label was \textless~1\% covered with LD classes.
A total of 73,552 image/label tiles were obtained, which were randomly divided into train (70\%), validation (20\%) and test (10\%) sets.

\subsection{Semantic Segmentation Model}

\subsubsection{Architecture}

\begin{figure*}[t]
  \centering
  \includegraphics[width=1.0\textwidth]{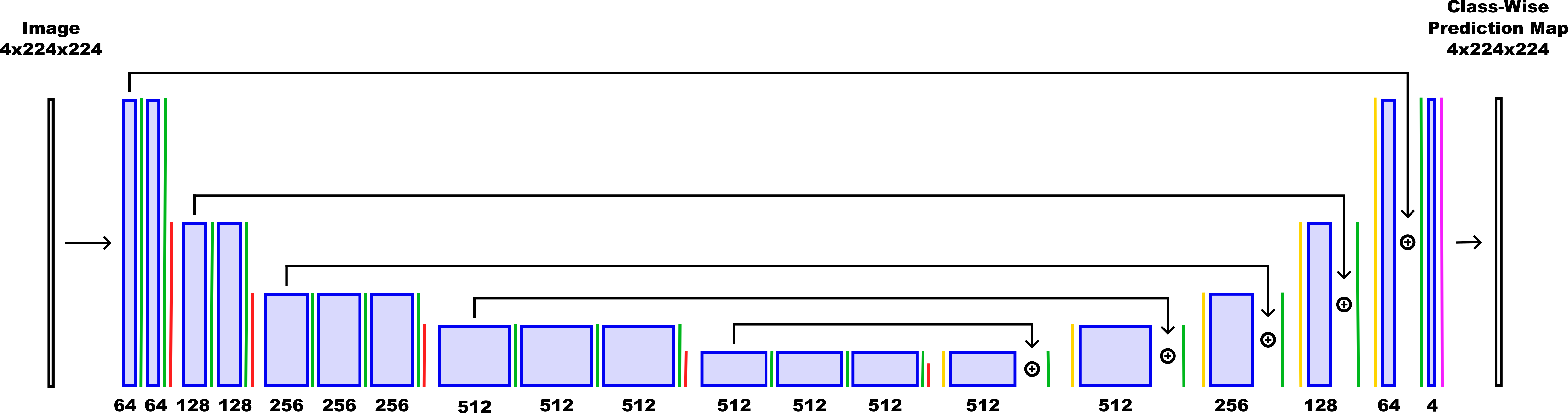}
  \caption{The VGGNet-16 architecture takes a $4\times224\times224$ image as input and outputs a $4\times224\times224$ per-class prediction map, from which a final $1\times224\times224$ label map is obtained using the argmax operation. Blue blocks denote $3\times3$ convolutional layers, with labels indicating the number of channels. Green lines denote a batch normalization layer followed by a ReLU activation function, red lines denote $2\times2$ max pooling layers, yellow lines denote bilinear up-sampling by a factor of 2, $\oplus$ denotes summation operation, and the magenta line denotes a hyperbolic tangent activation function.}
  \label{fig:vggnet}
\end{figure*}

For semantic segmentation, an encoder-decoder architecture is utilized, featuring the VGGNet encoder \cite{simonyan2015} and a modified U-Net decoder \cite{unet2015}. 
This segmentation model (illustrated in Fig.~\ref{fig:vggnet}) is chosen for its established efficacy in remote sensing segmentation tasks \cite{tsenov2023}.

The VGGNet encoder consists of a series of convolutional layers with small receptive fields (typically 3x3), followed by batch normalization layers, rectified linear unit (ReLU) activation functions, and max-pooling layers for down-sampling the feature maps while preserving essential information. 
These layers encode high-level semantic information from the input image. 
The modified U-Net decoder also features convolutional layers followed by batch normalization layers and ReLU activation functions.
However, it differs from the encoder by up-sampling the feature maps using bilinear interpolation to increase spatial resolution, while also incorporating skip connections from the encoder to facilitate better feature propagation and localization.
Finally, the model's last layer is a hyperbolic tangent function, producing a $4\times224\times224$ per-class prediction map, from which the $1\times224\times224$ label map is obtained using the argmax operation.

\subsubsection{Training}

The semantic segmentation model is trained by minimizing a composite loss function, which combines a weighted categorical cross entropy loss and a weighted Jaccard loss (intersection over union):
\begin{equation}
L(y, \hat{y}) = L_{CE}(y, \hat{y}) + L_J(y, \hat{y}),
\label{eqn:vgg_loss}
\end{equation}
where $\hat{y}$ is the softmax output of the per-class predicted label map produced by the model, and $y$ is the ground-truth label map.
Specifically, the weighted categorical cross entropy loss is defined as
\begin{equation}
    L_{CE}(y, \hat{y}) = -\frac{1}{N}\sum_{i=0}^N \sum_{c=0}^3 w_c \cdot y_c^i \cdot \log(\hat{y}_c^i) ,
    \label{eqn:ce}
\end{equation}
and the weighted Jaccard loss as
\begin{equation}
    L_J(y, \hat{y}) = \frac{1}{N}\sum_{i=0}^N\sum_{c=0}^3 w_c \cdot \left(1 - \frac{y_c^i \cdot \hat{y}_c^i}{y_c^i + \hat{y}_c^i -y_c^i \cdot \hat{y}_c^i}\right),
\end{equation}
where $y_c^i$ is a binary indicator if the true label of pixel $i$ belongs to class $c$, $\hat{y}_c^i$ is the predicted probability that pixel $i$ belongs to class $c$, $N$ is the number of pixels in an image, and $w_c$ is the weighting parameter.
Here, $w_c$ is chosen as $w_c=0.3$ for $c=1,2,3$ and $w_0=0.1$, corresponding to the inverse of the class imbalance, such that the background pixels constitute roughly 90\% of the images, and linear disturbances cover approximately 10\%.

The loss function (Eqn.~\ref{eqn:vgg_loss}) is minimized using the Adam optimizer \cite{adam} over two training phases: initially, for 100 epochs with a learning rate of 0.001, followed by an additional 100 epochs with a reduced learning rate of 0.0001. 
Images are normalized using z-score normalization where each pixel $x$ is transformed to $\frac{x-\mu}{\sigma}$, where $\mu$ and $\sigma$ represent the sample mean and standard deviation calculated per band over the entire tile set.
Tiles are grouped into batches of size 4 during training, and various augmentations, including random rotation, flipping, and zoom cropping, are applied to the images to prevent the model from overfitting the train set.

In addition, per-class predictions on the validation set are computed after each epoch and processed using argmax to create a $1\times224\times224$ predicted label map. 
Subsequently, the macro average mean Intersection over Union (mIoU) is computed between the predicted label map and the groundtruth label map as follows:
\begin{equation}
    \overline{\text{mIoU}} = \frac{1}{4}\sum_{c=0}^3 \text{mIoU}_c \label{eqn:macro_iou}
\end{equation}
where the per-class mIoU is
\begin{equation}
    \text{mIoU}_c = \frac{\text{TP}_c}{\text{TP}_c+\text{FP}_c+\text{FN}_c}. \label{eqn:per_class_iou}
\end{equation}
Here, the true positives $\text{TP}_c$ are the number of correctly predicted pixels for class $c$, the false positives $\text{FP}_c$ are the number of pixels incorrectly predicted as class $c$, and the false negatives $\text{FN}_c$ are the number of pixels incorrectly predicted as not class $c$ (including all other classes).
To avoid overfitting, if the validation $\overline{\text{mIoU}}$ does not decrease over 15 consecutive epochs, the model training progresses to the second training phase, or halted entirely if already in the second phase. 
Finally, the model weights selected are those corresponding to the lowest validation $\overline{\text{mIoU}}$ observed during the training phase.

\subsubsection{Testing}\label{sec:testing}

To properly evaluate the segmentation performance, the model is tested on a separate testing set that was not used during training. 
The evaluation metrics include per-class precision $P_c$ and recall $R_c$, defined as follows:
\begin{align}
P_c &= \frac{\text{TP}_c}{\text{TP}_c + \text{FP}_c}, \\
R_c &= \frac{\text{TP}_c}{\text{TP}_c + \text{FN}_c}.
\end{align}
Additionally, the per-class F1-score is computed as:
\begin{equation}
F1_c = 2 \cdot \frac{P_c \cdot R_c}{P_c + R_c},
\end{equation}
and the per-class mIoU is calculated according to Eqn.~\ref{eqn:per_class_iou}. 
Both a macro average,
\begin{equation}
    \overline{M} = \frac{1}{4}\sum_{c=0}^3 M_c
\end{equation}
and a weighted average,
\begin{equation}
    M = \frac{1}{4}\sum_{c=0}^3 p_c \cdot M_c
\end{equation}
is computed on each per-class metric $M_c$, where $p_c$ is the pixel proportion of class $c$.

\subsubsection{Implementation Details}

Training and testing of the model were conducted using the \textit{TensorFlow} \cite{tensorflow} Python library within \textit{Docker} \cite{Docker} containers on a ThinkStation P3 Tower with Core i9-13900 Processor (up to 5.20 GHz), an NVIDIA RTX A2000 12GB GDDR6 graphics card, and 64 GB of DDR5-4400MHz memory.
The model took approximately 72 minutes per epoch to train, and 5 minutes to test.

\section{Results}
\subsection{Segmentation Performance Overview}
\renewcommand{\arraystretch}{1.3} % Increase the row height by 50%
\begin{table}
\begin{center}
\caption{Per class testing performance.}
\label{table:seg_stats}
\begin{tabular}{| c || c | c | c | c | c |}
\hline
\textbf{Class} & \textbf{Precision} & \textbf{Recall} & \textbf{F1 Score} & \textbf{mIoU} & \textbf{Support}\\
$c$ & $P_c$ & $R_c$ & $F1_c$ & $\text{mIoU}_c$ & \\
\hline
\hline
Background & 0.966 & 0.972 & 0.969 & 0.940 & $3.5 \times 10^8$\\
\hline
Pipeline & 0.579 & 0.477 & 0.523 & 0.354 & $4.3\times10^6$\\
\hline
Road & 0.789 & 0.764 & 0.776 & 0.635 & $6.4\times10^6$\\
\hline 
Cutline & 0.517 & 0.478 & 0.497 & 0.331 & $1.5\times10^7$\\
\hline
\hline
Weighted & 0.940 & 0.943 & 0.942 & 0.903 & - \\
Avg. $M$ & & & & & \\
\hline
Macro & 0.713 & 0.673 & 0.692 & 0.565 & - \\
Avg. $\overline{M}$ & & & & & \\
\hline
\end{tabular}
\end{center}
\end{table}

\begin{figure}[ht]
  \centering
  \includegraphics[width=0.5\textwidth]{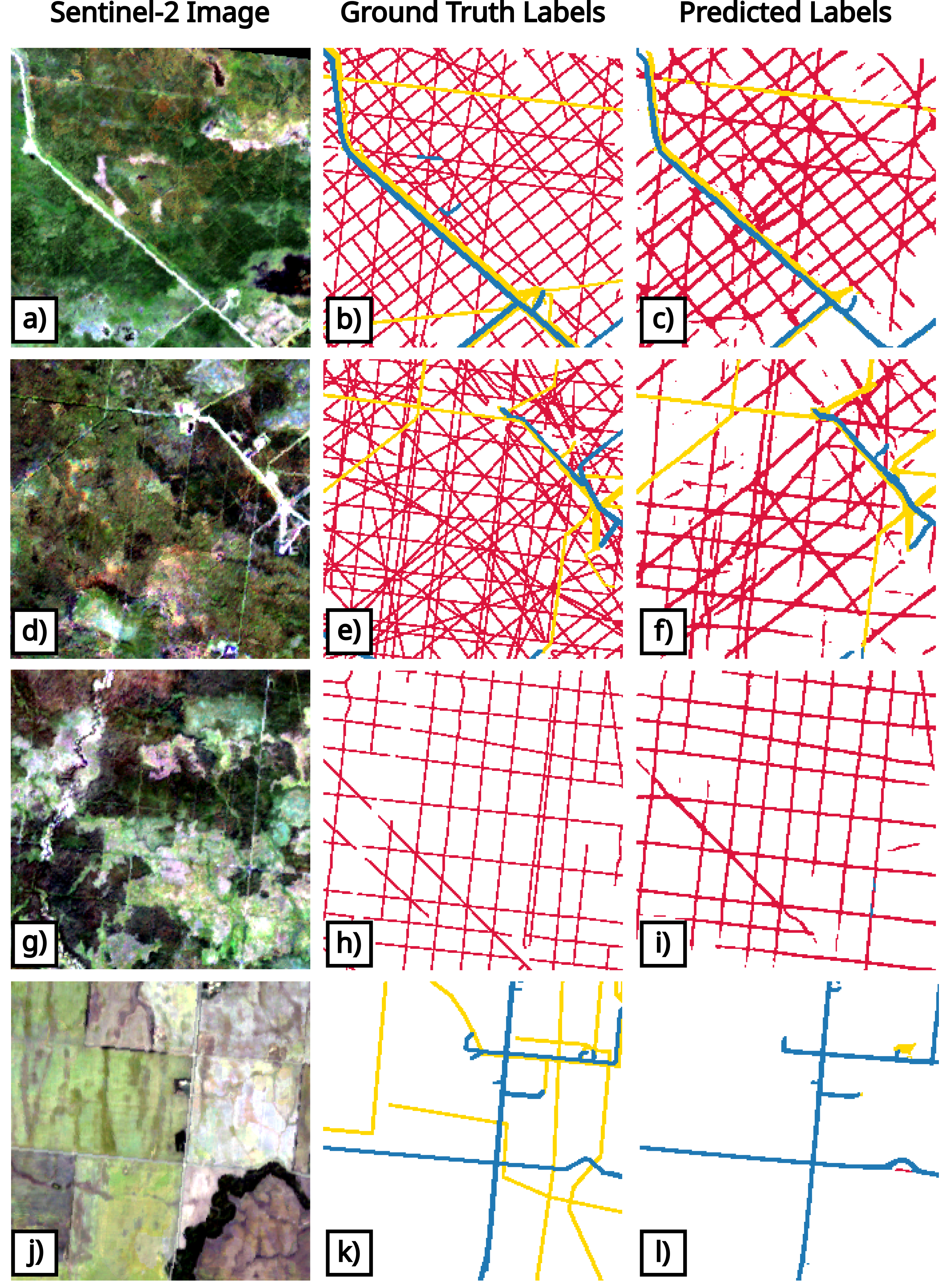}
  \caption{Visualization of Sentinel-2 imagery, ground-truth labels, and predicted labels for three test tiles. Each row represents a different test tile, which was not seen during training. The first column shows the original Sentinel-2 imagery. The second column displays the corresponding ground-truth labels, and the third column shows the predicted labels generated by the model. In the ground-truth and predicted label images, yellow denotes pipelines, blue denotes roads, and red denotes cutlines.}
  \label{fig:test_chips}
\end{figure}

The testing performance of the segmentation model is summarized in Table~\ref{table:seg_stats}. 
The best-segmented linear disturbance (LD) class is roads, with an F1 score of 0.776 and an mIoU of 0.635.
The model's strong performance on this class can be attributed to roads generally being wider than pipelines and cutlines, making them visually more prominent in the S2 imagery. 
Additionally, the presence of paved roads in this class enhances their detectability and makes segmentation easier.
An example of a testing tile that produces a well-segmented road is shown in Fig.~\ref{fig:test_chips}. 
The road predicted by the model, denoted in blue in Fig.~\ref{fig:test_chips}c), closely matches the ground-truth road, also in blue, in Fig.~\ref{fig:test_chips}b). 
As expected, the corresponding S2 image in Fig.~\ref{fig:test_chips}a) shows a thick, bright, paved road that is clearly visible in the imagery.

Nonetheless, despite their larger size, the support for the road class reported in Table~\ref{table:seg_stats} is relatively low, comprising about 1.7\% of total pixels and roughly 25\% of linear disturbance pixels in the testing set.
Moreover, the road class comprises 14 distinct sub-classes, such as pavement roads, gravel roads, and vegetated road edges.
The model's ability to generalize well across these varied sub-classes and with only a small amount of support is noteworthy.

The next best-segmented LD class is pipelines, achieving an F1 score of 0.523 and an mIoU of 0.354. 
The decrease in segmentation performance for this class compared to roads can be attributed to three main factors.
Firstly, the support for the pipeline class, as reported in Table~\ref{table:seg_stats}, shows that pipelines occur even less frequently in the dataset compared to roads, comprising about 1.1\% of total pixels and roughly 17\% of linear disturbance pixels in the testing set.
The reduced data for this class makes learning and generalizing pipelines more challenging.

Secondly, and most importantly, many of the pipelines are not visible in the imagery. 
They can only be detected by the swath of vegetation removed above them, which may make them appear as roads or cutlines, or even result in them being indistinguishable as disturbances.
Finally, many pipelines run parallel to roads (as shown in Fig.~\ref{fig:cutline_raster}), and the 10m resolution of the S2 imagery may not capture the boundary between these features clearly. 
Often, this results in parallel roads and pipelines appearing as wide roads in the imagery.
In addition, the rasterization process of the HF vector labels can further blur or adjust this boundary making it challenging to accurately segment the pipelines.

Despite pipelines being a much more challenging task to segment, the model can still produce valuable predictions as shown in Fig.~\ref{fig:test_chips}.
The pipelines predicted by the model, denoted by yellow in Fig.~\ref{fig:test_chips}f), closely resemble the ground-truth pipelines, also in yellow in Fig.~\ref{fig:test_chips}e), with only a small amount of confusion with the neighbouring road in blue.
This is an impressive result, considering the model's ability to discern that the pipeline is running parallel to the road, even though they appear as a single wide feature in the corresponding S2 imagery shown in Fig.~\ref{fig:test_chips}d).
Thus, despite the quantitative metrics of F1 score and mIoU indicating mediocre performance for the pipeline class, the model qualitatively produces valuable predictions about the locations of linear features in the S2 imagery.

Finally, the cutline class is the most challenging linear disturbance (LD) class to segment, achieving an F1 score of 0.497 and an mIoU of 0.331. 
The reduced performance for this class can be attributed to several factors: the artificial thickening of the cutline labels during rasterization, the lower contrast of cutlines in imagery compared to more prominent features like paved roads, the overall diversity within the cutline class, and the thinner nature of cutlines compared to most roads and pipelines. 
Many cutlines have a thickness of only 1-3 pixels, making them nearly impossible to discern in the 10m S2 imagery. 
Consequently, a significant portion of cutlines are not be captured at all or may be lost in the complex structure of land cover in this landscape in relation to S2 resolution. 
Furthermore, cutlines often appear in diverse landscapes, such as forests or rocky terrain, adding to the segmentation difficulty. 

Similar to pipelines, the cutline class presents challenges in segmentation; however the model can still generate valuable predictions, as demonstrated in Fig.~\ref{fig:test_chips}.
The predicted cutlines, marked in red in Fig.~\ref{fig:test_chips}i), show a reasonable correspondence with the ground-truth cutlines, also denoted by red in Fig.~\ref{fig:test_chips}g).
While some cutlines are missing or inaccurately predicted, this is a notable achievement given that many cutlines cannot be detected visually in the corresponding S2 imagery in Fig.~\ref{fig:test_chips}g), and the rest are extremely challenging to distinguish amid the complex structure of land cover in this landscape in relation to the S2 resolution.

\begin{figure}[ht]
  \centering
  \includegraphics[width=0.45\textwidth]{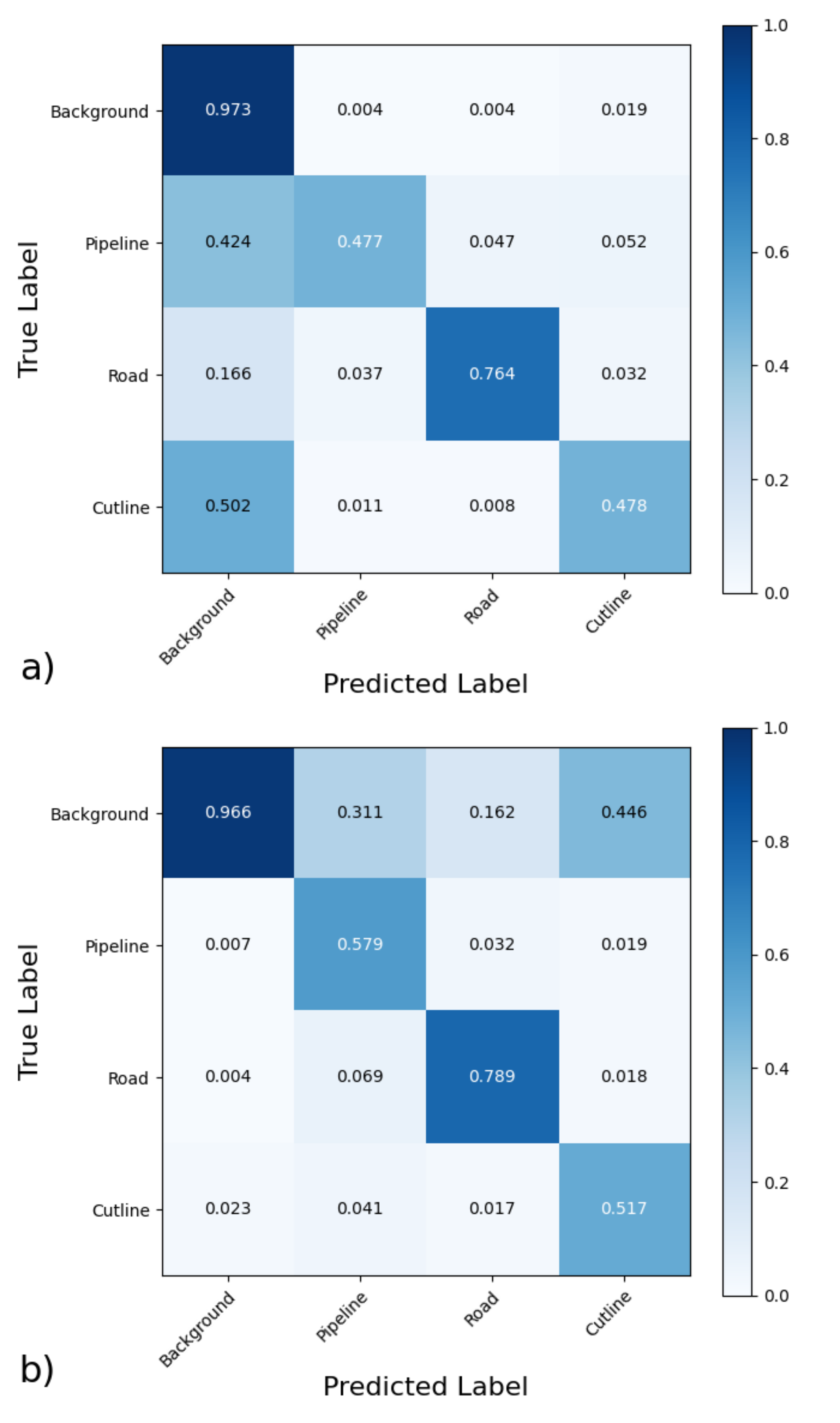}
  \caption{Confusion matrices normalized: \textbf{a)} by rows to show recall along the diagonal and \textbf{b)} by columns to show precision along the diagonal.}
  \label{fig:confusion_matrices}
\end{figure}

To further analyze the segmentation of LDs, a confusion matrix is used to display the number of correct and incorrect pixel predictions broken down by each class.
For ease of interpretation, the pixel counts are normalized across the rows (true labels) in Fig.~\ref{fig:confusion_matrices}a) to show recall along the diagonal of the confusion matrix, and normalized across the columns (predicted labels) in Fig.~\ref{fig:confusion_matrices}b) to show precision along the diagonal.

For all LD classes, the first column of the recall confusion matrix in Fig.~\ref{fig:confusion_matrices}a) indicates that the majority of the missed LD pixels are falsely predicted as background pixels.
This misclassification generally occurs when LDs are completely missed by the model or, in rare cases, when the predicted width of the feature is narrower than the ground truth.
Such outcome is unexpected, given that many LDs, such as thin cutlines, are scarcely visible in the coarser-resolution 10m S2 imagery, and can be easily overlooked or undetectable by the segmentation model.
This issue is exemplified by the testing tiles displayed in Fig.~\ref{fig:test_chips}, where numerous ground-truth cutlines (red) visible in Fig.~\ref{fig:test_chips}e), cannot be detected visually in the S2 imagery shown in Fig.~\ref{fig:test_chips}d).
Consequently, these features are not predicted by the model, as illustrated by the predicted labels in Fig.~\ref{fig:test_chips}f), which entirely omit several of the cutlines.
Another example is the pipeline (yellow) at the bottom of the ground-truth label in Fig.~\ref{fig:test_chips}b), which is absent in the S2 imagery in Fig.~\ref{fig:test_chips}a), and also omitted by the models prediction shown in Fig.~\ref{fig:test_chips}c).

In addition to completely missing linear features and misclassifying them as background, Fig.~\ref{fig:confusion_matrices}a) shows that a small portion of the recall error (approximately 0.03-0.05) arises from confusion between LD classes. 
Specifically, pipeline pixels are sometimes predicted as road or cutline pixels, and road pixels are sometimes predicted as pipeline or cutline pixels.
Quantitative evidence of this confusion is illustrated by the testing tiles shown in Fig.~\ref{fig:test_chips}, where a road (blue) located in the middle right of the ground-truth label in Fig.~\ref{fig:test_chips}e) is incorrectly predicted as a pipeline (yellow) in Fig.~\ref{fig:test_chips}f).

Similar to the recall confusion matrix in Fig.~\ref{fig:confusion_matrices}a), which shows that the primary source of recall error is LD labels being incorrectly predicted as background, the precision confusion matrix in Fig.~\ref{fig:confusion_matrices}b) indicates that the main source of precision error arises from background pixels being falsely predicted as LD labels. 
There are several reasons for this misclassification.
One primary reason is when the model predicts the width of a feature to be wider than the ground truth, as seen in the predicted labels in Fig.~\ref{fig:test_chips}c), where the cutlines appear thicker than the ground truth cutlines in Fig.~\ref{fig:test_chips}b). 
Additionally, errors can result from the model misclassifying pixels in a random or linear pattern, especially when identifying landscape features that resemble LDs. 
This is evident in the middle right side of Fig.~\ref{fig:test_chips}i), where a diagonal cutline is predicted but does not exist in the ground-truth labels shown in Fig.~\ref{fig:test_chips}h).
Finally, the model may extrapolate portions of fractured or unconnected LD features, as demonstrated in the left side of Fig.~\ref{fig:test_chips}i), where fragmented cutlines observed in the ground-truth labels in Fig.~\ref{fig:confusion_matrices}h) are incorrectly filled in by the model. 
Like the recall error, Fig.~\ref{fig:confusion_matrices}b) shows that confusion between LD classes contributes only a small portion of the precision error.

For all linear disturbance classes, the model's precision is higher than its recall, as shown in both Table~\ref{table:seg_stats} and Fig.~\ref{fig:confusion_matrices}. 
This indicates that the model is more likely to miss disturbance pixels altogether rather than mislabel background pixels as disturbances or confuse different disturbance types. 
This outcome is expected, given that the primary source of error in detecting disturbances stems from the S2 resolution in relation to narrow disturbances, leading to their omission by the model.
This issue is less pronounced for thicker road disturbances, as reflected by the model's recall being much closer to its precision for this class.

In general, the model performs well at segmenting the LD dataset as a whole, with a weighted average F1 score of 0.942 and a weighted average mIoU of 0.903 reported in Table~\ref{table:seg_stats}.
However, these high values are skewed due to the large class imbalance in the dataset, with well-segmented background pixels contributing to over 93\% of the mIoU value.
For a more balanced assessment of the model's performance, the macro average F1 score of 0.692 and macro average mIoU of 0.565 are much better predictors.

\subsection{Spatially Distributed Segmentation Performance}

\begin{figure*}[t]
  \centering
  \includegraphics[width=1.0\textwidth]{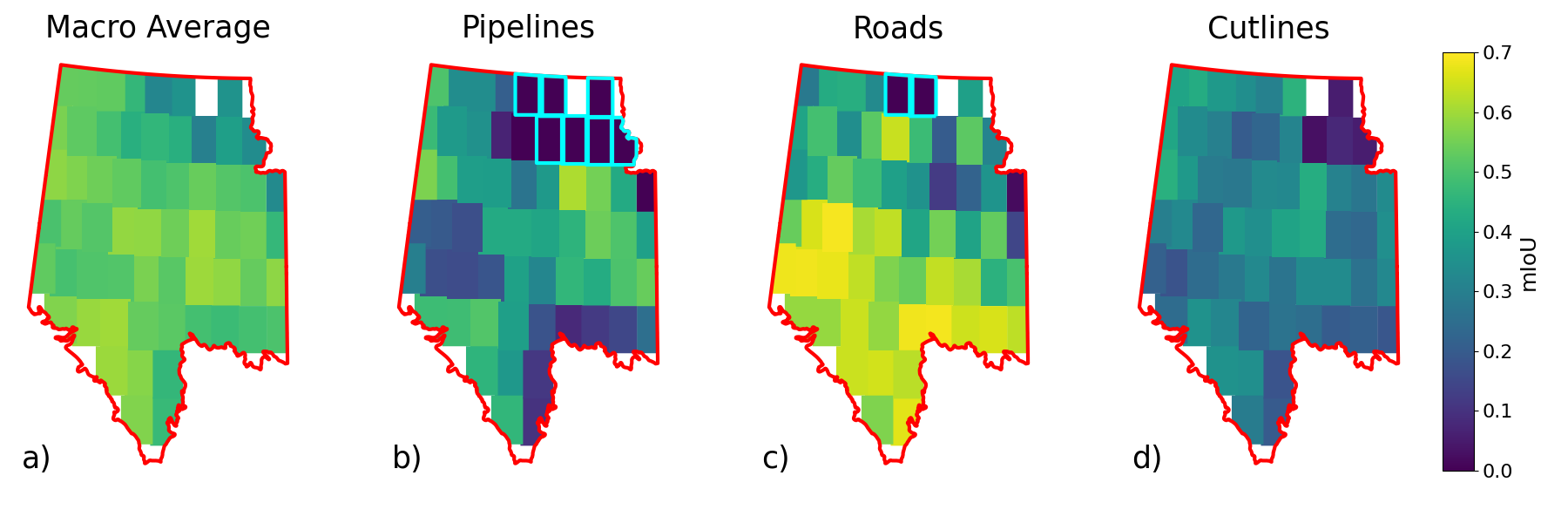}
  \caption{The mean intersection over union (mIoU) computed on testing tiles over sub-regions of the total study area (denoted by red outline), with the mIoU value indicated by color. Blue outlines indicate sub-regions that contain no support for the respective class. \textbf{a)} Macro average mIoU, \textbf{b)} mIoU of the pipeline class, \textbf{c)} mIoU of the road class, and \textbf{d)} mIoU of the cutline class.}
  \label{fig:spatial_mIoU}
\end{figure*}

In Fig.~\ref{fig:spatial_mIoU}, the model's performance is broken down spatially by dividing the study area into various sub-regions, and computing the mean intersection over union (mIoU) over testing tiles located in each sub-region. 
The four subplots correspond to the (a) macro averaged value $\overline{\text{mIoU}}$ and the per class value $\text{mIoU}_c$ of the (b) pipeline, (c) road, and (d) cutline classes. 
The color of each sub-region denotes the mIoU value as indicated by the colorbar, and blue outlines denote sub-regions with an mIoU value that contain no support for the respective class.

The spatially distributed $\overline{\text{mIoU}}$ displayed in Fig.~\ref{fig:spatial_mIoU}a) suggests that the model performs relatively uniformly across the entire study area. 
However, a slight performance degradation is observed near the top right corner of the study area, corresponding to a region with very sparse amounts of linear disturbances (LDs), making it challenging for the model to learn and generalize over the landscape. 
This observation is particularly corroborated by all $\text{mIoU}_c$ values in Fig.~\ref{fig:spatial_mIoU}b-d), which also show performance degradation in the top right of the study area. 
Since much of this region contains areas with no pipelines or roads (as indicated by the blue outlined sub-regions), any false positives predicted for the underrepresented classes would result in an $\text{mIoU}_c$ values near zero.

The spatially distributed $\text{mIoU}_c$ of the pipeline and road classes, displayed in Fig.~\ref{fig:spatial_mIoU}b) and c) respectively, show a less uniform distribution of accuracy across the study area. 
Interestingly, aside from the top right region of the study area where both classes have lower support resulting in lower mIoU values, generally the regions where pipelines are segmented poorly correspond to regions where roads are segmented accurately. 
In particular, road predictions are more accurate on average in the lower half of the study area, where the Boreal Plains landscape approaches the Prairie ecozone, and many roads are either paved and easily identifiable, are formed in grids that provide a predictable structure to identify, or divide stretches of agriculture which are also easy to distinguish in the imagery. 
An example of a testing tile from this sub-region is included in Fig.~\ref{fig:test_chips}j), where roads are separating square plots of land making them visually distinct in the S2 imagery, so the predicted labels denoted by blue in Fig.~\ref{fig:test_chips}l) correspond very closely to the ground-truth labels also denoted in blue in Fig.~\ref{fig:test_chips}k).

The reason for the poor segmentation of pipelines in these sub-regions can also be explained by the test tile in Fig.~\ref{fig:test_chips}j). 
Many pipelines in urban areas are actually buried underground and are entirely obscured from the S2 imagery, resulting in poor pipeline segmentation. 
This is evident in Fig.~\ref{fig:test_chips}, where the ground-truth pipelines denoted by yellow in Fig.~\ref{fig:test_chips}k) are visually undetectable in the Sentinel-2 imagery in Fig.~\ref{fig:test_chips}j) so the model does a poor job at segmenting them in Fig.~\ref{fig:test_chips}l). 
Thus, despite the model’s mediocre performance on pipelines as indicated in Table~\ref{table:seg_stats}, much of the error results from these types of scenarios, where underground pipelines do not actually appear as linear disturbances in the landscape and possibly should not be considered as such.
This suggests that the model’s actual performance is better than what the performance metrics indicate.

Finally, the spatially distributed $\text{mIoU}_c$ of the cutline class, displayed in Fig.~\ref{fig:spatial_mIoU}d), is fairly uniform across the study area. 
Given that cutlines are the thinnest disturbance and the most challenging to identify, the detection of these features is dominated by the resolution of the S2 imagery and less sensitive to the specific sub-region in which the feature is located.
Overall, it is noteworthy that the model demonstrates effective generalization across the study area, despite the significant variability in the landscape and the diversity within each linear disturbance class.

\section{Conclusions}

This study developed and evaluated an automated algorithm for accurately extracting linear disturbances across the Boreal and Taiga Plains ecozones of Alberta, which includes several boreal caribou herd ranges, using deep learning techniques and S2 imagery. 
Specifically, the research focused on a model that combines a VGGNet encoder with a modified U-Net decoder for pixel-level multi-class semantic segmentation of linear disturbances in forest settings, utilizing the HF dataset as ground-truth labels. 
The model demonstrated high effectiveness in segmenting wider features, such as roads, and moderate accuracy when segmenting narrower features like pipelines and cutlines. 
Despite being uniformly accurate across the study area, the model generally showed greater accuracy in segmenting roads near residential areas compared to forested areas, highlighting this study's unique contribution over previous research that primarily focused on urban regions.

Even though linear disturbance classes often appear similar to one another or run parallel, the model was able to segment them with minimal confusion between classes. 
One primary error arose from the model entirely missing some disturbances. 
This error led to a decrease in recall, primarily due to either thin cutlines not being captured in the 10m S2 imagery or underground pipelines in residential areas that do not appear in the imagery. 
Consequently, the recall is heavily influenced by these scenarios, where it is seemingly impossible for the model to detect the features in the imagery.

Another primary error involved the model identifying linear disturbances but predicting the features to be wider than they exist in the ground truth dataset. 
This error led to a decrease in precision. 
Nonetheless, qualitative results show that the model still produces useful linear disturbance maps, and the quantitative results may overestimate the error. 
Moreover, if simply detecting the presence of linear disturbances is more critical than their exact width, the model's utility is enhanced.

This research demonstrates the potential of a cost-effective method using deep learning architectures and S2 data to maintain current and accurate maps of linear disturbances in dynamic forest areas, supporting caribou conservation efforts. 
Building on the methods developed in this study, large areas could be mapped frequently, potentially creating a comprehensive national linear disturbance database, aiding in decision-making for caribou habitat conservation.

We recommend that future work explore different image preprocessing techniques that might enhance the visibility of linear features. 
Since the original labels were produced at a much higher resolution, less visible features pose challenges in both training and testing coarse-resolution imagery such as S2. Accordingly, super-resolution approaches to increase the resolution of S2 images using higher-resolution ground-truth datasets may be a promising area for future research in linear disturbance detection. 
Alternatively, directly training segmentation models on high-resolution satellite data could result in much more accurate detection, albeit with a trade-off in cost due to the need to purchase commercial datasets.

\bibliographystyle{plain}
\bibliography{references}  %%% Uncomment this line and comment out the ``thebibliography'' section below to use the external .bib file (using bibtex) .





% \end{thebibliography}

\end{document}